\documentclass{article}

\PassOptionsToPackage{numbers, compress}{natbib}

\usepackage[final]{neurips_2024}

\usepackage[utf8]{inputenc} %
\usepackage[T1]{fontenc}    %
\usepackage[colorlinks,linkcolor=red]{hyperref}       %
\usepackage{url}            %
\usepackage{booktabs}       %
\usepackage{amsfonts}       %
\usepackage{nicefrac}       %
\usepackage{microtype}      %
\usepackage{xcolor}         %
\usepackage{graphicx}
\usepackage{amsmath}
\usepackage{multicol}
\usepackage{multirow}
\usepackage{wrapfig}
\usepackage{booktabs}
\usepackage{bbding}

\makeatletter
\newcommand\figcaption{\def\@captype{figure}\caption}
\newcommand\tabcaption{\def\@captype{table}\caption}
\makeatother

\makeatletter
\def\thanks#1{\protected@xdef\@thanks{\@thanks
        \protect\footnotetext{#1}}}
\makeatother

\title{Zero-Shot Scene Reconstruction from Single Images with Deep Prior Assembly}

\author{Junsheng Zhou$^1$\quad\;\; \textbf{Yu-Shen Liu}$^{1\dagger}$\thanks{$^\dagger$The corresponding author is Yu-Shen Liu.}\quad\;\; \textbf{Zhizhong Han$^2$}\\
School of Software, Tsinghua University, Beijing, China$^1$ \\ 
Department of Computer Science, Wayne State University, Detroit, USA$^2$\\
{\tt\small zhou-js24@mails.tsinghua.edu.cn\hspace
{3mm}liuyushen@tsinghua.edu.cn\hspace
{3mm}h312h@wayne.edu}
}

\begin{document}

\maketitle

\begin{abstract}
Large language and vision models have been leading a revolution in visual computing. By greatly scaling up sizes of data and model parameters, the large models %
learn deep priors which lead to remarkable performance in various tasks. In this work, we present deep prior assembly, a novel framework that assembles diverse deep priors from large models for scene reconstruction from single images in a zero-shot manner. We show that this challenging task can be done without extra knowledge but just simply generalizing one deep prior in one sub-task. To this end, we introduce novel methods related to poses, scales, and occlusion parsing which are keys to enable deep priors to work together in a robust way.
Deep prior assembly does not require any 3D or 2D data-driven training in the task and demonstrates superior performance in generalizing priors to open-world scenes. 
We conduct evaluations on various datasets, and report analysis, numerical and visual comparisons with the latest methods to show our superiority. Project page: \url{https://junshengzhou.github.io/DeepPriorAssembly}.
\end{abstract}

\section{Introduction}
Reconstructing scenes from images is a vital task in 3D computer vision and computer graphics. It bridges the gap between the 2D images that can be easily captured by phone cameras and the 3D geometries of scenes for various real-world applications, e.g., autonomous driving, augmented/virtual reality and robotics.
Reconstructing scenes from multi-view images \cite{wang2021neus,yumonosdf} is well-explored to recover 3D geometries with multi-view consistency and camera poses. However, reconstructing a scene from a single RGB image is still challenging, which is extremely difficult due to inadequate information. Recent works \cite{dahnert2021panoptic, nie2020total3dunderstanding} try to solve this task as a reconstruction problem which leverages neural networks with an encoder-decoder architecture to draw supervisions from pairs of images and 3D ground truth geometries and layouts. Nevertheless, due to the limited amount of image-scene pairs, these methods struggle to generalize to out-of-distribution images in open world. 

Large language and vision models have been extensively studied in the past few years, which revolutionized neural language processing \cite{touvron2023llama, brown2020language}, 2D/3D representation learning \cite{fang2023eva, zhou2023uni3d} and content generation \cite{rombach2022high, jun2023shap}, etc. By greatly scaling up sizes of training samples and model parameters, large models show brilliant capabilities and remarkable performance. However, they are limited in a specific task, 
%with a specific modality, which 
which limits their capability in high level perception tasks. Driven by the observation, we raise an interesting question: can we assemble series of deep priors from large models, which are experts with different modalities in different tasks, to solve an extremely challenging task that none of them can accomplish alone?

\begin{figure}[t]
    \centering
    \includegraphics[width= \linewidth]{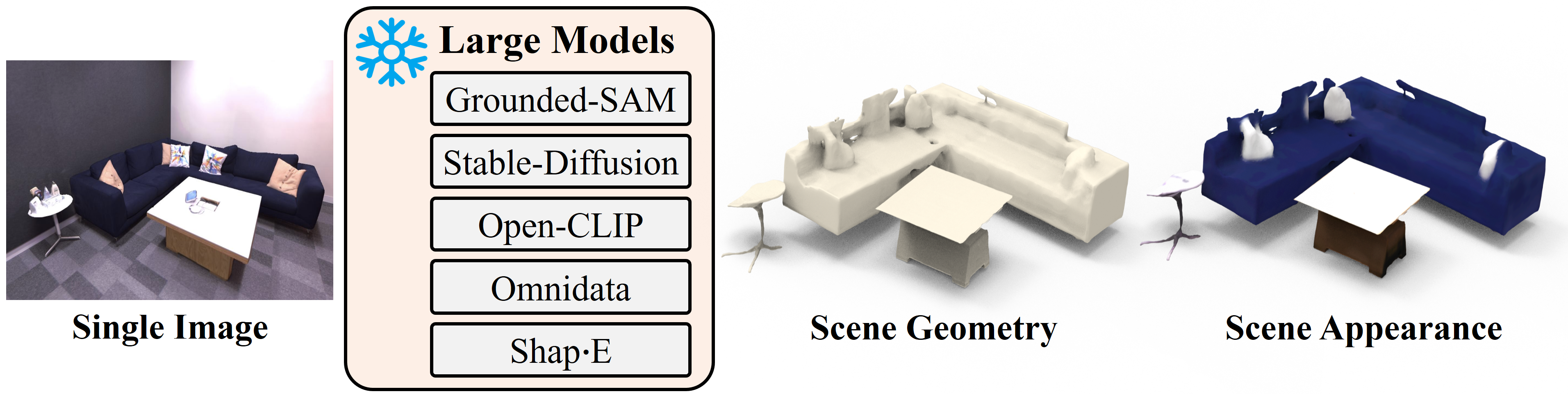}
    \caption{\textbf{An illustration of our work.} We assemble diverse deep priors from large models with frozen parameters for scene reconstruction from single images in a zero-shot manner.}
    \label{fig:enter-label}
    \vspace{-0.5cm}
\end{figure}

In this work, we propose \textit{deep prior assembly}, a novel framework which assembles diverse deep priors from large models for scene reconstruction from single images in a zero-shot manner.
We rethink this task from a new perspective, and decompose it into a set of sub-tasks instead of seeking to a data-driven solution. We narrow down the responsibility of each deep prior on a sub-task that it is good at, and introduce novel methods related to poses, scales, and occlusion parsing to enable deep priors to work together in a robust way.

Specifically, we first detect and segment the instances in the input image with Grounded-SAM \cite{kirillov2023segment, liu2023grounding}, which is a variation of Segment-Anything Model \cite{kirillov2023segment}. For the segmented instances that are often corrupted due to occlusions or of low resolution, we leverage Stable-Diffusion \cite{rombach2022high} to enhance and inpaint images containing the segmented instances. However, the Stable-Diffusion often produces some predictions which drift away from the input instances and do not conform to the original appearance. To solve this issue, we introduce to use Open-CLIP \cite{radford2021learning, ilharco_gabriel_2021_5143773} to filter out the bad samples and select the ones matching the input instance most. We then generate the 3D models for each instance with Shap$\cdot$E \cite{jun2023shap} using the amended instance images as input. Additionally, we estimate the depth of the origin image with Omnidata \cite{eftekhar2021omnidata} as the 3D scene geometry prior. 
To recover a layout consistent to the image, 
we propose an approach to optimize the location, orientation and size for each 3D instance to fit it to the estimated segmentation masks and the depths.  

Deep prior assembly merely generalizes deep priors and does not require additional data-driven training for extra prior knowledge. Our evaluations on various open-world scenes show our capability of reconstructing diverse objects and recovering plausible layout merely from a single view angle. 
Our main contributions can be summarized as follows.
\begin{itemize}
    \item We propose deep prior assembly, a flexible framework that assembles diverse deep priors from large models together for reconstructing scenes from single images in a zero-shot manner.
    \item We introduce a novel approach to optimizing the location, orientation and scale of instances by matching them with both 2D and 3D supervision. 
    \item We evaluate deep prior assembly for generating diverse open-world 3D scenes, and show our superiority over the state-of-the-art supervised methods.
\end{itemize}

\section{Related Work}

\subsection{Large Models in Different Modalities}
Recently, it has been drawing significant attention on scaling up deep models for much more powerful representations and higher performances with different modalities (e.g. NLP, 2D vision). Starting from NLP, recent works in scaling up pre-trained language models~\citep{brown2020language, liu2019roberta, raffel2020exploring} have largely revolutionized natural language processing.
Some researches translates the progress from language to 2D vision~\citep{radford2021learning, dosovitskiy2020image, bao2021beit, he2022masked, fang2023eva} and 3D vision \cite{zhou2023uni3d} via model and data scaling. 

Except for the large foundation models which focus on producing large-scale representations for language, 2D images or 3D point clouds, some researches explore large models for specific tasks (e.g. text-to-image generation \cite{rombach2022high}, image segmentation \cite{kirillov2023segment}, 3D analysis \cite{xu2023pointllm,zhou20223d,Zhou2023VP2P,li2024LDI,li2023neaf} and 3D object generation \cite{udiff,wen20223d,ma2023geodream,zhou2024diffgs}) and have shown remarkable performance. Stable Diffusion trains a large model of latent diffusion models and achieves commercially available 2D content generation effects. Segment Anything Model (SAM) \cite{kirillov2023segment} revolutionize the field of image segmentation by training models with large-scale annotated data. Omnidata \cite{eftekhar2021omnidata} trains the large depth estimation model with various data sources for bringing robust 3D awareness to pure RGB images. In the 3D domain, the recent works Point$\cdot$E \cite{nichol2022point} and Shap$\cdot$E \cite{jun2023shap} collect millions of 3D objects to train large 3D models for generating 3D geometries from rendering-style images or texts. In this work, we aim at leveraging the powerful capabilities of the large models in different modalities and different domains to solve a challenging task, i.e. scene generation from single images, by assembling deep priors together.

\subsection{Scene Reconstruction from Images}
Recovering the underlying 3D surfaces of scenes from images \cite{wang2021neus,zhang2023fast,huang2023neusurf,han2024binocular,nie2023learning} or point clouds \cite{zhou2024cap,Zhou2022CAP-UDF,zhou2023levelset,jin2023multi,ma2023towards,zhou2024fast,jin2024music,takeshi2024multipull} is a long-standing task in 3D computer vision. Most of the previous works focus on the multi-view reconstruction with the input dense images captured around the scene. Classic multi-view stereo methods \cite{agrawal2001probabilistic, broadhurst2001probabilistic, bleyer2011patchmatch} mainly represent the scene by estimating depths for dense images with feature matching. Inspired by NeRF \cite{mildenhall2020nerf} which performs volume rendering for scene representation, a series of works \cite{wang2021neus, oechsle2021unisurf, yariv2021volume, wang2022neuris, zhang2024gspull, zhang2024learning} introduce the neural implicit surface reconstruction by learning signed distance fields \cite{park2019deepsdf} or occupancy fields \cite{mescheder2019occupancy} for scenes from multi-view images. NeuRIS \cite{wang2022neuris} proposes to use normal priors for indoor scene reconstruction, and MonoSDF \cite{yumonosdf} further introduces monocular depth cues for improving scene geometries.

\section{Method}

\begin{figure*}[t]
    \centering
    \includegraphics[width=\linewidth]{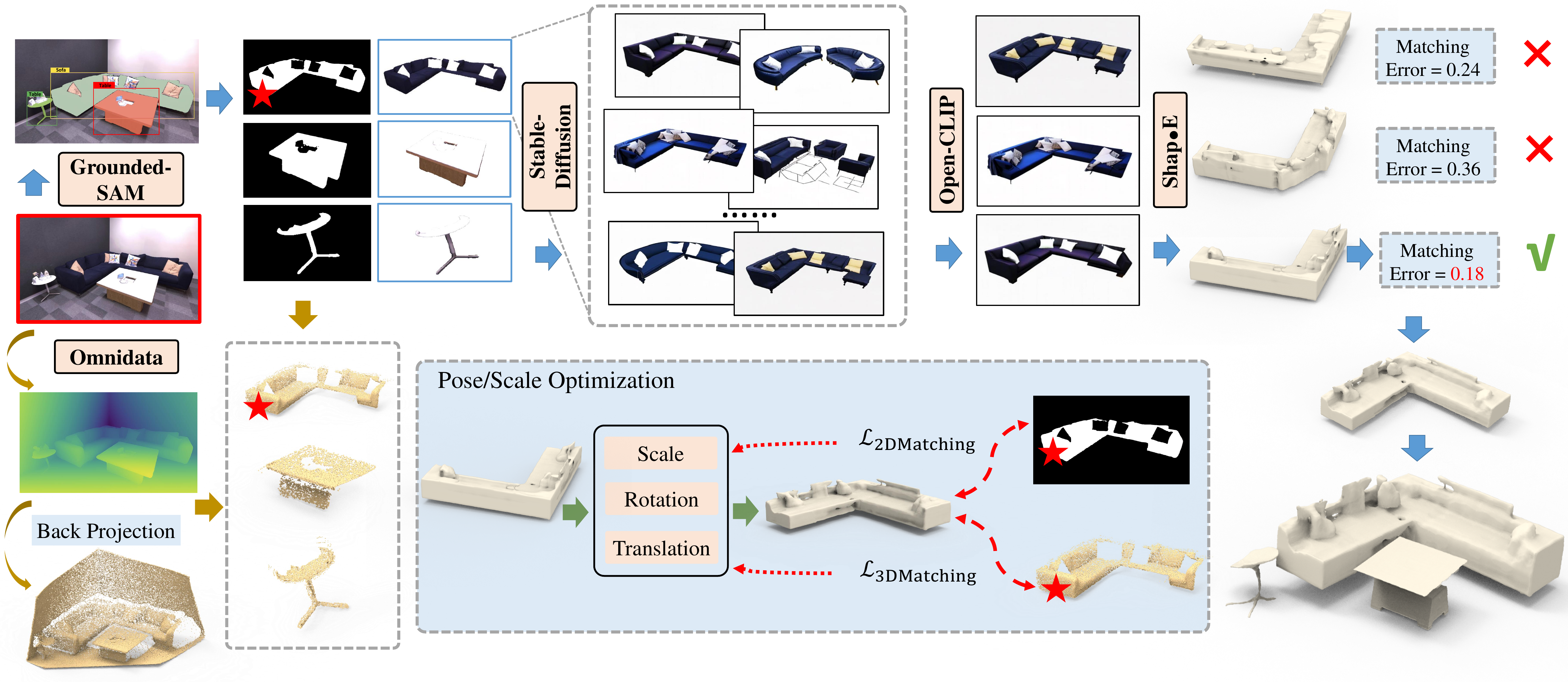}
    \vspace{-0.6cm}
    \caption{\textbf{The overview of deep prior assembly.} Given a single image of a 3D scene, we detect the instances and segment them with Grounded-SAM. After normalizing the size and center for the instances, we attempt to amend the quality of the instance images by enhancing and inpainting them. Here, we take a sofa in the image for example. Leveraging the Stable-Diffusion model, we generate a set of candidate images through image-to-image generation, with additional guidance from a text prompt of the instance category predicted by Grounded-SAM. We then filter out the bad generation samples with Open-CLIP by evaluating the cosine similarity between the generated instances and original one. After that, we generate multiple 3D model proposals for this instance with Shap$\cdot$E from the Top-$K$ generated instance images. Additionally, we estimate the depth of the origin input image with Omnidata as a 3D geometry prior. To estimate the layout, we propose an approach to optimize the location, orientation and scale for each 3D proposal by matching it with the estimated segmentation masks and the depths (the \textcolor{red}{$\star$} for the example sofa). Finally, we choose the 3D model proposal with minimal matching error as the final prediction of this instance, and the final scene is generated by combining the generated 3D models for all detected instances.}
    \label{fig:overview}
    \vspace{-0.3cm}
    
\end{figure*}

\textbf{Overview.}
The overview of deep prior assembly is shown in Fig. \ref{fig:overview}. 
We will start from an introduction of the task decomposition in Sec. \ref{sec.3.1} and then present the pipeline for solving each of the decomposed sub-tasks using a deep prior from a specific large model in Sec. \ref{sec.3.2}. Finally, we introduce an optimization-based approach for layout estimation in Sec. \ref{sec.3.3}. 

\subsection{Task Decomposition}
\label{sec.3.1}
Revealing 3D scene geometries from a single image is an extremely challenging task duo to  limited context and supervisions. 
Instead of using a data-driven strategy to learn priors~\cite{nie2020total3dunderstanding, dahnert2021panoptic}, we reformulate this task from a new perspective. We decompose it into a set of sub-tasks, each of which can be done using one deep prior without a need of learning extra knowledge.
More specifically, we can progressively resolve the task by:

\begin{itemize}
    \item[1)] First, performing detection and segmentation on the input image to acquire the segmentation images, masks and category labels for all detected instances.
    \item[2)] Amending instance images through enhancing and inpainting to improve the image qualities.
    \item[3)] Generating a set of 3D model proposals for each instance from 2D segmented images.
    \item[4)] Estimating the layout by predicting the location, rotation, and scale for each 3D proposal to put them to the correct position of the 3D scene.
    \item[5)] Producing a scene reconstruction by applying the estimated layout and shape poses with reconstructed instances.
\end{itemize}

\subsection{Assembling Large Models}
\label{sec.3.2}
Inspired by the remarkable performances of recent large models, we propose to assign an expert large model in each sub-task, which maximizes their abilities for modeling a scene in a zero-shot manner.

\noindent\textbf{Detect and Segment 2D instances.}
To reveal a scene $S$ from a single image $I$, we first detect the instances in $I$ and separate multiple objects into single instances. In this way, we can reconstruct a scene at shape level, which simplifies the task.

\noindent\textit{Mask R-CNN vs. SAM vs. Grounded SAM.} Detecting and segmenting images have been widely explored in the past few years. Mask R-CNN \cite{he2017mask} is widely adopted as a popular and robust backbone. However, the performance of Mask R-CNN does not generalize well since it is only trained under a relative small dataset. Recently, the large SAM~\cite{kirillov2023segment} have shown promising segmentation accuracy by scaling up parameters and using more training samples, nevertheless, it only predicts fine-grained masks but with few semantic concepts. Thus, we adopt Grounded-SAM, which is an improved version of SAM by introducing Grounding-DINO \cite{liu2023grounding} as an open-set detector and using SAM to jointly predict detection boxes, segmentation masks and category labels for each instance, formulated as:
\begin{equation}
    \{m_i, c_i, d_i\}_{i=1}^N = \mathrm{GroundedSAM}(I), \; \text{where} \; d_i > \sigma
\end{equation}

$\{m_i, c_i, d_i\}$ includes the predicted mask $m_i$, category $c_i$ and the detection confidence score $d_i$ for the $i$-th instance, and $N$ is the number of instances in this scene. We only keep the predictions with a high confidence score larger than a threshold $\sigma$. The low confident instances often contain large occlusions or wrong category predictions.

\noindent\textbf{Enhance and Inpaint 2D instances.}
With the predicted masks $\{m_i\}_{i=1}^N$, we achieve the segmented instance images $\{t_i\}_{i=1}^N$ by masking the original input image $I$. We then normalize each instance image by centering it at the origin and normalizing its scale in $\{t_i\}_{i=1}^N$ to 0.6 of the max width or height of $I$. As shown in Fig. \ref{fig:overview}, the segmented instance images often suffer from occlusions or low-resolution of small instances. The low-quality images have a large negative impact on the followup 3D generation. Therefore, we propose to first improve the quality of instance images by enhancing and inpainting them with the large model Stable-Diffusion~\cite{rombach2022high}.

\begin{wrapfigure}[21]{r}{0.58\textwidth}
    \centering
    \vspace{-0.2cm}
    \includegraphics[width=0.96\linewidth]{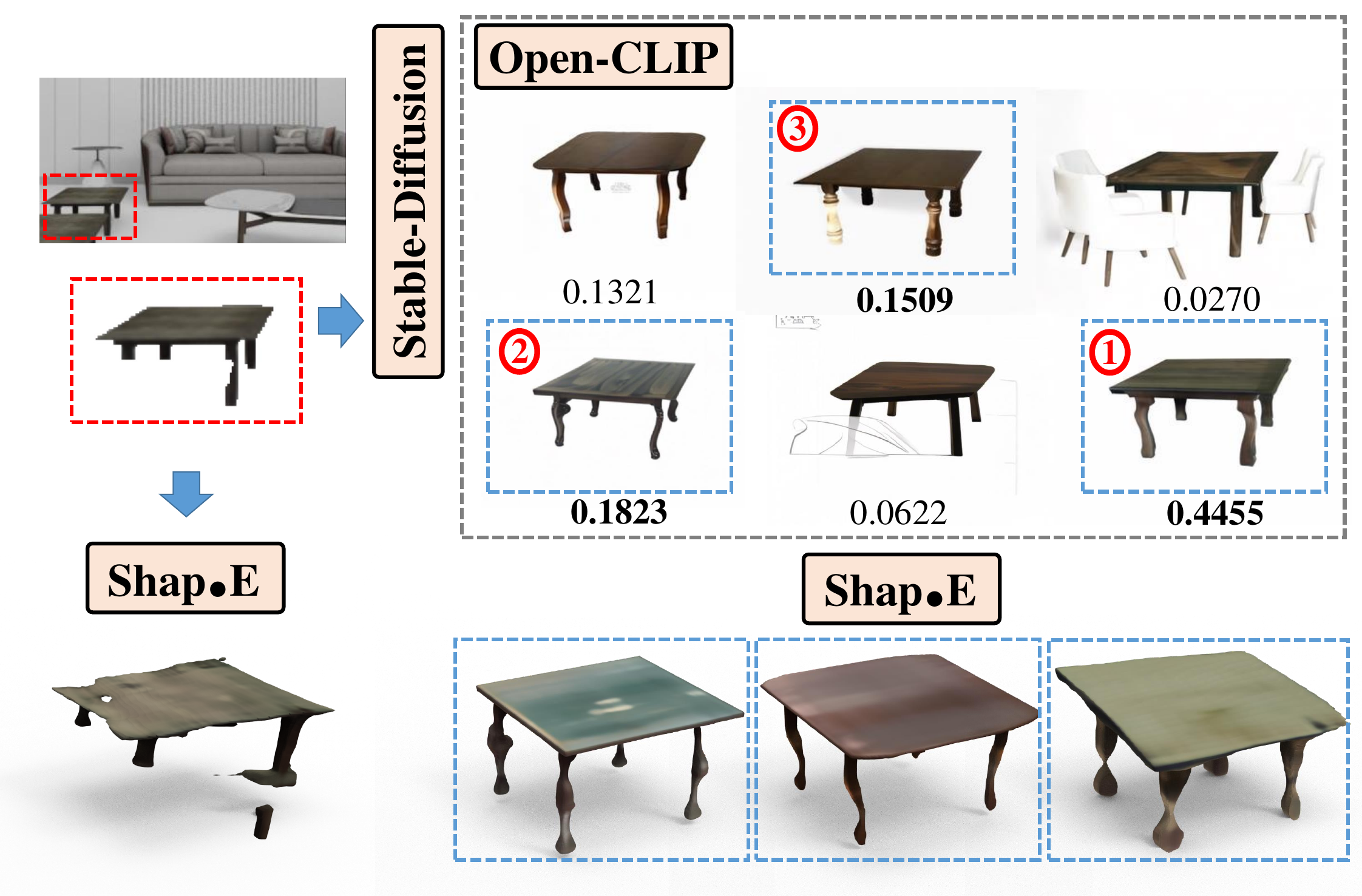}
    \vspace{-0.25cm}
    \caption{\textbf{Examples on the effect of our pipeline.} For the corrupted 2D instant segmented from the scene image, we leverage Stable-Diffusion to  produce $6$ amended generations. We then adopt Open-CLIP to filter out bad samples by judging the similarities and producing confidence scores for the generations, and keep the Top-$3$ generated images. The shape generations with Shap$\cdot$E from the amended images are significantly more complete and accurate than the one produced by the original corrupted image. }
    \label{fig:method_sd}
    \vspace{-0.3cm}
\end{wrapfigure}

Specifically, we adopt the image-to-image generation~\cite{meng2021sdedit} from Stable-Diffusion. We take the instance image $t_i$ as the initialization, and add noises on it and then subsequently denoise the noise corrupted image to increase the realism through the guidance of the text prompt description from the predicted categories $c_i$. We find the prompt template `High quality, authentic style \{category\}' works fine for most of the indoor instances. For other situations, we directly leverage the category as the prompt. We observe that Stable-Diffusion may produce some unreliable predictions which are `too creative' and can not faithfully improve the image quality but turn it into another image, as shown in Fig. \ref{fig:method_sd}.
To solve this issue, we generate multiple enhanced images $\{e_i^j\}_{j=1}^M$ for each instance image $t_i$ and filter out the bad generation samples with the approaches described next.

\noindent\textbf{Filter Out Bad Generation Samples.}
To filter out the bad generation samples produced by Stable-Diffusion and select the top $K$ enhanced images for the following 2D-to-3D generation, we propose to leverage the CLIP models as a judge to determine which generated images $\{e_i^j\}_{j=1}^M$ from Stable-Diffusion can conform to the original appearance of the instance $t_i$. Specifically, we adopt the large open-sourced CLIP \cite{radford2021learning} model Open-CLIP \cite{ilharco_gabriel_2021_5143773} as the implementation.

We use cosine similarities $\{z_i^j\}_{j=1}^M$ between the generated instance image $\{e_i^j\}_{j=1}^M$ and the original one $t_i$ as a metric for the selection, formulated as:
\begin{equation}
    z_i^j = \frac{f_{\theta}(e_i^j) \cdot f_{\theta}(t_i)}{\|f_{\theta}(e_i^j)\| \|f_{\theta}(t_i)\|},
\end{equation}
where $f_{\theta}$ is the frozen image encoder from Open-CLIP. We use the Top-$K$ generated instance images with the largest similarities to the original $t_i$ as the amended images. As shown in Fig. \ref{fig:method_sd}, Open-CLIP successfully filters out the bad generations.

\noindent\textbf{Generate 3D models from 2D instances.}
The object-level 3D generation from single images \cite{mescheder2019occupancy, wen20223d} is a well-explored task in 3D computer vision, however, most previous works show limited generation performance on open-world images. Shap$\cdot$E \cite{jun2023shap} is a large model for 3D generation by training a 3D hyper diffusion model with millions of non-public 3D objects. Therefore, we leverage Shap$\cdot$E to provide the deep prior to convert 2D instance images into 3D reconstructions. 
By this way, we obtain $K$ 3D reconstruction proposals $\{s_i^k\}_{k=1}^K$ for each 2D instance $t_i$ by employing Shap$\cdot$E on the top $K$ amended images.

\subsection{Recovering Scene Layout}
\label{sec.3.3}

The final step is to select the most accurate 3D model proposal $\hat{s_i}$ from the $K$ candidates $\{s_i^k\}_{k=1}^K$ and put it to the right position in a 3D scene to recover the scene layout in the input image. To achieve this, we propose a novel approach to optimize the location, orientation and size for each 3D proposal in $\{s_i^k\}_{k=1}^K$ by matching it with the estimated segmentation mask and the estimated depth. We further introduce a RANSAC-like solution for robust position optimizing. We select the reconstruction with the minimum matching error as the reconstruction $\hat{s_i}$ of $t_i$.

\begin{wrapfigure}[17]{r}{0.58\textwidth}
    \centering
    \vspace{-0.4cm}
    
    \includegraphics[width=\linewidth]{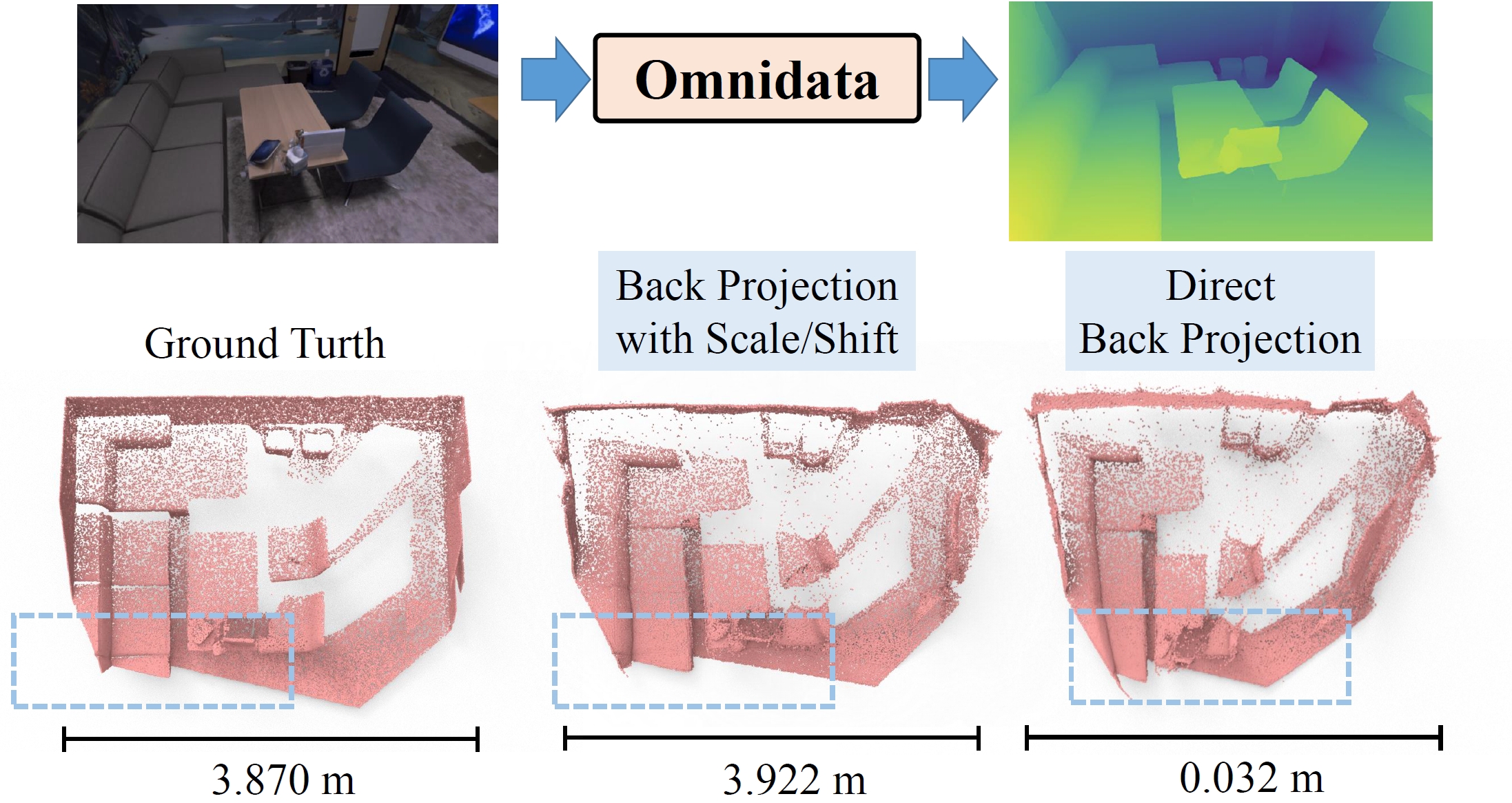}
    \vspace{-0.6cm}
    \caption{\textbf{Illustration of the depth transform.} The estimated depth maps from Omnidata is not scale-aware, resulting in scale inaccuracies and distortion in the back-projected depth point clouds. We achieve the accurate depth point cloud by first transforming the depth maps with the pre-solved scale and shift before back-projecting.}
    \label{fig:method_depth}
    \vspace{-0.3cm}
    
\end{wrapfigure}
\noindent\textbf{Depth Estimation.}
For more accurate layout estimations, we first estimate the depth map of the input image $I$ as a 3D geometry prior. We leverage the large model Omnidata \cite{eftekhar2021omnidata, kar20223d} as the depth estimator which is trained under a collected huge depth dataset \cite{eftekhar2021omnidata} containing 14-million indoor, outdoor, scene-level and object-level samples.

An issue here is that the depth $D$ estimated for the input image $I$ with Omnidata is not scale-aware,
which can not be directly used as supervisions.
We solve this problem by estimating the scale $h$ and shift $q$ of the predicted depth using one pair of predicted and real depth of a selected scene for each dataset.  Specifically, we leverage least-squares criterion \cite{eigen2014depth, ranftl2020towards} which has a closed-form solution to solve the depth scale and shift by matching the pair of predicted and real depth with a specific scene camera intrinsic parameter $\mathbb{C}_{K}$. After transforming $D$ with the scale $h$ and shift $q$, we can now back-project $D$ into the 3D space with camera intrinsic parameter $\mathbb{C}_{K}$, achieving a 3D scene depth point cloud. The example in Fig.~\ref{fig:method_depth} shows that the depth point cloud produced using the transformed depth maps precisely aligns with the ground truth scene. Alternatively, we can use metric depth estimation methods \cite{yang2024depth, yin2023metric3d,depth_anything_v2,ke2024repurposing}, which naturally handle depth scale and shift, to replace Omnidata and lessen the reliance on ground truth depth data. The depth point cloud $d_i$ for each instance $t_i$ is further achieved by masking the back-projected 3D point cloud.

\noindent\textbf{Pose/Scale Optimization.}
We further estimate the scale and pose of each 3D model proposal $s_i^k$ to put them into the right position in the 3D scene. We propose to solve this problem with an optimization-based approach on the location, rotation and size for $s_i^k$ per the mask $m_i$ from the Grounded-SAM and the depth point cloud $d_i$ from Omnidata. 

We model this problem as a 7-DoF shape registration task with 3-DoF of translation ($tx$, $ty$, $tz$), 3-DoF of rotation ($rx$, $ry$, $rz$) and one DoF of object scale ($v$). Specifically, we first sample a point cloud $p_i^k$ from the mesh of a 3D proposal $s_i^k$ and initialize a 7-DoF transform as a transformation function $f_{\phi}$ with learnable 7-DoF parameters $\phi$. We then project $p_i^k$ with $f_{\phi}$ to achieve the transformed prediction $\hat{p}_i^k$ by:
\begin{equation}
    \hat{p}_i^k = f_{\phi}(p_i^k).
\end{equation}
We obtain $\hat{p}_i^k$ by optimizing the 7-Dof parameters $\phi$ with supervisions. With the estimated depth $d_i$, we can draw the direct 3D matching supervision by minimizing the Chamfer Distance Loss between the transformed $\hat{p}_i^k$ and the depth points $d_i$. However, merely with the 3D matching constraint, the pose/scale optimization do not always converge stably since the predicted depth $d_i$ is usually with noises in complex scenes, which significantly affects the registration on shapes.

\begin{wrapfigure}[20]{r}{0.6\textwidth}
    \centering
    \vspace{-0.4cm}
    \includegraphics[width=\linewidth]{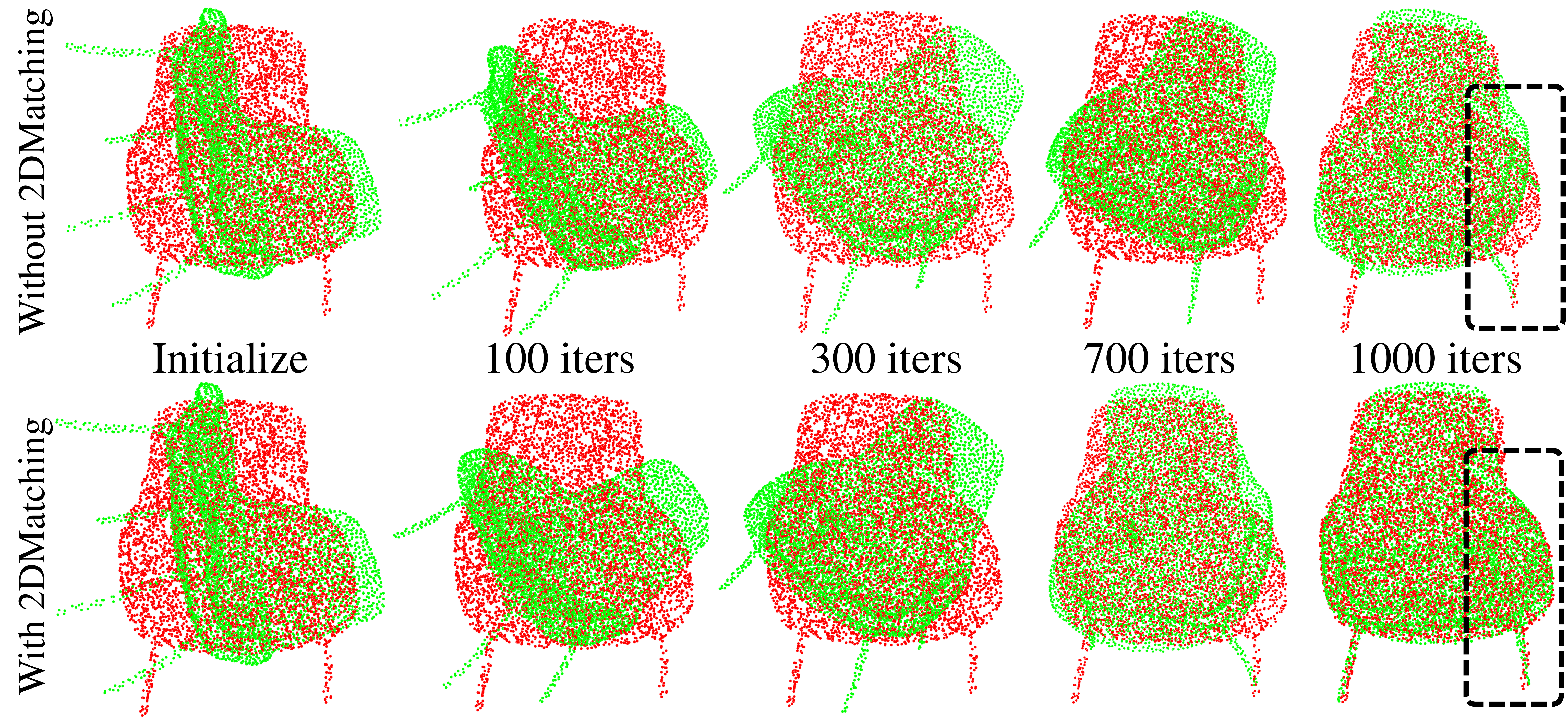}
    \caption{\textbf{Effect of the 2D Matching.} An example of optimizing the pose and scale for a chair. We visualize the optimization in 2D space. The red 2D points indicate the dense 2D point cloud sampled in the mask, which is the target. And the green 2D points donate the 2D projection of transformed 3D point clouds sampled from the generated shape of this chair instance. More robust registration is achieved with the proposed 2D matching constraint. The total 1,000 iterations take $9.2$ seconds on a single 3090 GPU.}
    \label{fig:method_2dmatching}
    \vspace{-0.3cm}
    
\end{wrapfigure}

To resolve this issue, we get inspirations from \cite{chen2021unsupervised} to leverage the mask information predicted by Stable-Diffusion as an extra matching supervision in 2D space. Specifically, we project the  transformed 3D point cloud $\hat{p}_i^k$ to the 2D space with the camera intrinsic parameters $\mathbb{C}_{K}$, resulting in a 2D point cloud $\tilde{p}_i^k$. Meanwhile, we form another 2D point cloud $\tilde{m}_i$ from the mask $m_i$ by randomly sampling dense 2D points on the occupied region of the mask $m_i$. We then use the 2D matching constraint to minimize the 2D Chamfer Distance between $\tilde{p}_i^k$ and $\tilde{m}_i$. We illustrate the effect of 2D Matching constraint with an optimization example in Fig. \ref{fig:method_2dmatching}. The final loss for pose/scale optimization of 3D reconstruction $s_i^k$ is then formulated as:
\begin{equation}
\label{eq.matching}
    \mathcal{L} = \mathcal{L}_{CD}^{3D}(d_i, \hat{p}_i^k) + \mathcal{L}_{CD}^{2D}(\tilde{p}_i^k, \tilde{m}_i).
\end{equation}

\noindent\textbf{Robust RANSAC-like Solution.}
With the optimization-based 7-DoF registration, we are now able to put the generated 3D object proposals into the 3D scene. However, if the mis-registration is quite large, especially in the rotation, the optimization may be trapped in a local optimum and fail to produce accurate registrations. We further introduce a RANSAC-like solution to enhance the robustness of pose/scale optimization. Specifically, we repeat the optimization $r$ times with randomly initialized rotation matrices for $f_{\phi}$ each time. The final transform for the 3D proposal $s_i^k$ is selected as the one with minimum matching loss in Eq. (\ref{eq.matching}) among $r$ optimal optimizations, and we define the matching error $w_i^k$ of $s_i^k$ as the minimal matching loss.

We repeat the above procedure for each one of the $K$ 3D proposals $\{s_i^k\}_{k=1}^K$. We select the 3D proposal with the minimum $w_i^k$ as the final reconstruction $\hat{s_i}$ of $t_i$.
The final scene generation from the single image $I$ is achieved by combining the transformed $\{\hat{s_i}\}_{i=1}^N$ together.  

\begin{table*}[!tb]\small
\setlength{\abovecaptionskip}{1mm}
\centering

\resizebox{\textwidth}{!}{
\begin{tabular}{l|c|c|c|c|c|c|c|c|c}
\toprule
\multirow{1}*{} & \multicolumn{3}{c|}{3D-Front} & \multicolumn{3}{c|}{BlendSwap} & \multicolumn{3}{c}{Replica}\\
\cline{2-10}

Method  & CDL1-S  $ \downarrow$ & CDL1 $ \downarrow$ & F-Score$\uparrow$ & CDL1-S $ \downarrow$ & CDL1 $ \downarrow$ & F-Score$\uparrow$ & CDL1-S $ \downarrow$ & CDL1 $ \downarrow$ & F-Score$\uparrow$ \\ 
\midrule
Mesh R-CNN \cite{gkioxari2019mesh}  & 0.449 & 0.471 & 23.90 & 0.265 & 0.406 & 21.87 & 0.268 & 0.408 & 25.42 \\
Total3D \cite{nie2020total3dunderstanding} & 0.198 & 0.520 & 18.44 &  0.133 & 0.400 & 26.93 & 0.390 & 0.780 & 24.01\\
PanoRecon \cite{dahnert2021panoptic} & 0.120 & \textbf{0.125} & 31.94 & 0.355 & 0.417 & 17.11 & 0.326 & 0.440 & 17.13\\
\midrule
Ours & \textbf{0.109} & 0.134 & \textbf{35.67} & \textbf{0.106} & \textbf{0.089} & \textbf{73.19} & \textbf{0.113} & \textbf{0.110} & \textbf{70.48} \\
\bottomrule
\end{tabular}}
\caption{Comparisons on scene reconstruction from single images. Lower is better for CDL1 (i.e., Chamfer Distance), higher is better for F-Score. CDL1-S is the single-direction Chamfer Distance from the generated objects to the ground truth meshes.}
\label{table:main}
\end{table*}

\begin{figure*}[t]
    \centering
    \includegraphics[width=\linewidth]{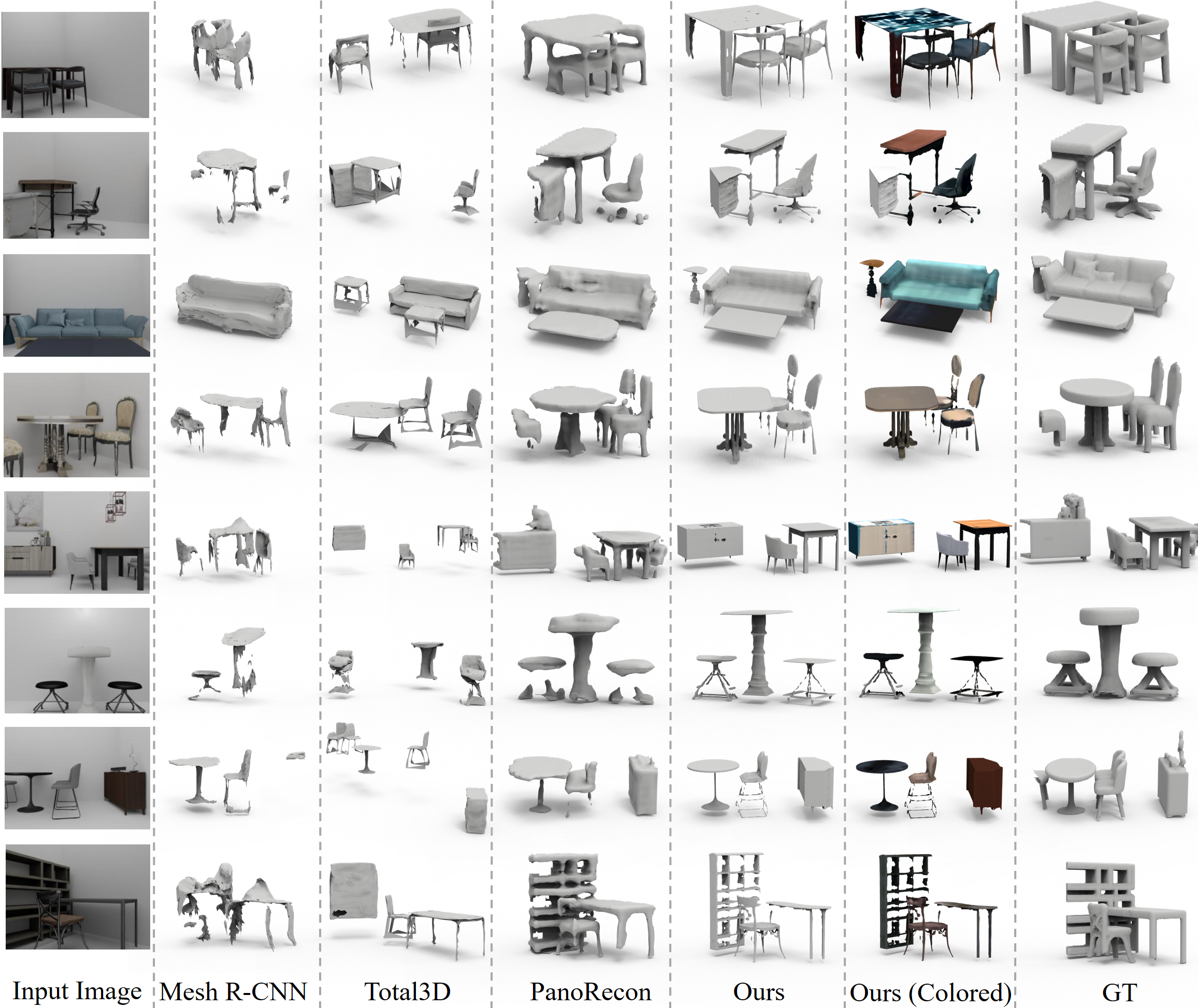}
    \vspace{-0.5cm}
    
    \caption{Comparisons on scene reconstruction from single images under the 3D-Front dataset.}
    \label{fig:3dfront}
    \vspace{-0.6cm}
    
\end{figure*}

\begin{figure*}[t]
    \centering
    \includegraphics[width=\linewidth]{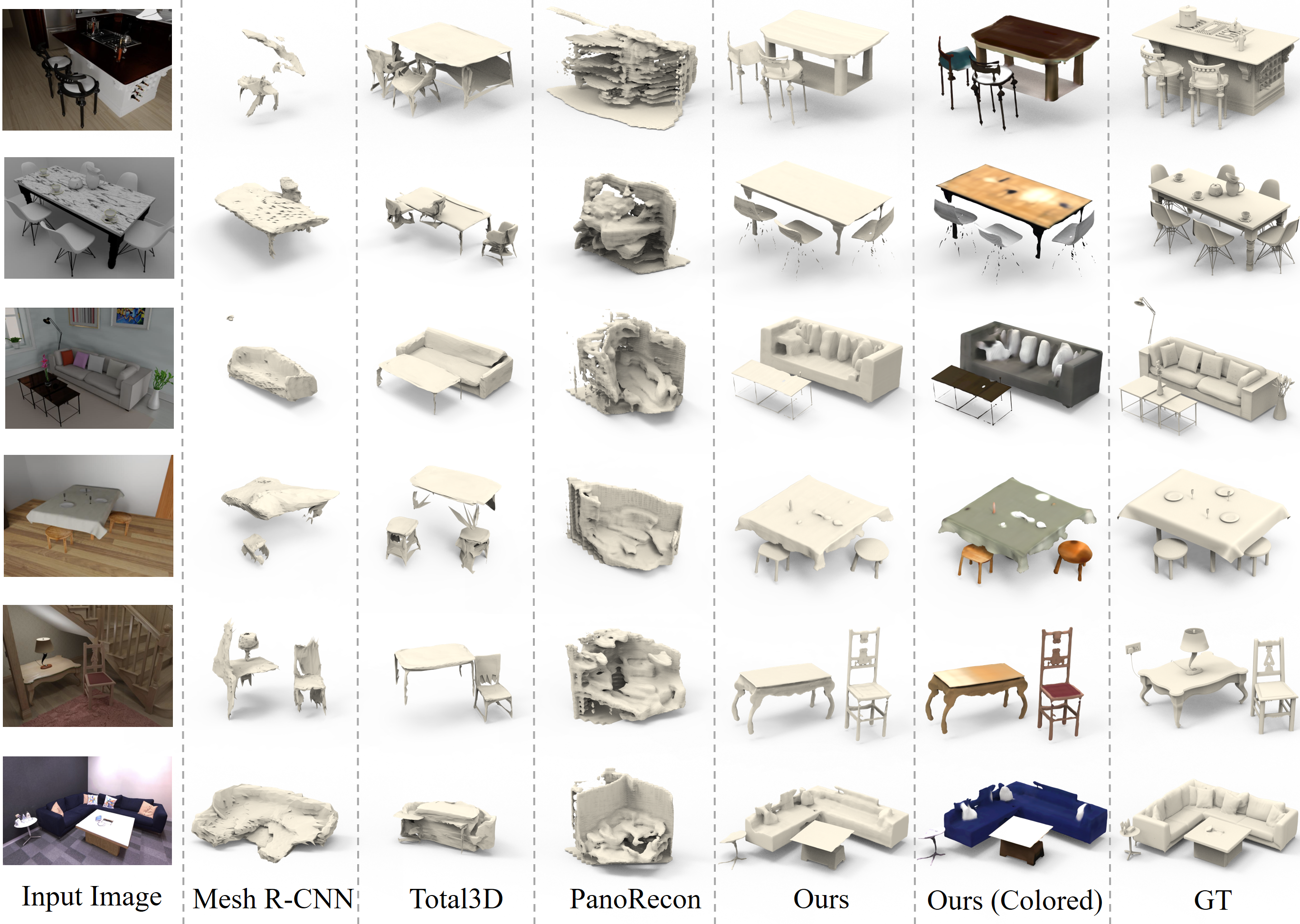}
    \vspace{-0.5cm}
    \caption{Comparisons on scene reconstruction from single images under Replica and BlendSwap dataset.}
    \label{fig:replica}
    \vspace{-0.7cm}
    
\end{figure*}

\vspace{-0.3cm}
\section{Experiments}

\subsection{Setup}

\noindent\textbf{Implement Details.} The number $M$ of samples generated by Stable-Diffusion for each instance is set to $6$, where we select the Top $K=3$ samples with Open-CLIP. The pose/scale optimization is repeated for $r=10$ times for each instance with RANSAC-like solution. 

\noindent\textbf{Datasets.}
We evaluate deep prior assembly under four widely-used 3D scene reconstruction benchmarks 3D-Front \cite{fu20213d}, Replica \cite{straub2019replica}, BlendSwap \cite{azinovic2022neural} and ScanNet \cite{dai2017scannet}. 

3D-Front \cite{fu20213d} is a synthetic 3D dataset of indoor 3D scenes. We adopt the data pre-processed by PanoRecon \cite{dahnert2021panoptic} and randomly select 1,000 scene images from the test set as the single-image dataset. Note that all the images are captured parallel to the ground with camera locations at 0.75m height above the floor in the 3D-Front dataset. We follow PanoRecon to achieve the corresponding ground truth mesh for each image by only keeping the geometry at the same view and cull anything outside of the view frustum. 

The Replica \cite{straub2019replica} dataset is an indoor scene dataset which contains 8 scanned 3D indoor scene with highly photo-realistic 3D indoor scene reconstruction at both room and flat scale. We adopt the pre-processed data provided by MonoSDF \cite{yumonosdf} and sample one image for each scene as the single-image dataset. The ground truth meshes are obtained with the same way as 3D-Front.

The BlendSwap \cite{azinovic2022neural} dataset is a high-fidelity synthesis 3D scene dataset collected by Neural-RGBD \cite{azinovic2022neural}, containing 9 scenes with complex geometries. We collect single-view images and corresponding ground truth meshes with the same way as Replica dataset.

The ScanNet \cite{dai2017scannet} dataset is a real-word 3D scene dataset captured by RGB-D cameras. We select 7 scenes from the test set of ScanNet and sample one image from each scene as the input. 

\noindent\textbf{Baselines.}
We mainly compare our method with the state-of-the-art methods in scene reconstruction from single images, i.e., Mesh R-CNN \cite{gkioxari2019mesh}, Total3D \cite{nie2020total3dunderstanding} and PanoRecon \cite{dahnert2021panoptic}. Note that all these methods are data-driven methods and trained under 3D datasets with ground truth 3D annotations, while our method solves the task in a zero-shot manner. This means that we do not require any data-driven training on any 3D or 2D datasets, which is a much more flexible and general solution for the single image reconstruction task. We direct evaluate these methods with the official codes and the pre-trained models for numerical and visual comparisons.

\noindent\textbf{Metrics.}
We use Chamfer Distance and F-Score with the default threshold of 0.1 following \cite{nie2020total3dunderstanding, peng2021shape} as metrics. Since Mesh R-CNN and Total3D only predicts the 3D objects and do not generate the backgrounds (e.g. wall and floor), we further report the single-direction Chamfer Distance from the generated objects to the ground truth meshes, i.e., CDL1-S, to only evaluate the accuracy of generated objects. Note that Total3D can generate the scene layout which can roughly represent the background, however, we find that Total3D generates layouts with large errors on all the three datasets. Therefore we do not keep the layout of Total3D for evaluation. While we achieve the background points by back-projecting the segmented background depth maps. We sample 10k points on the ground truth meshes and the generated scenes of each methods for evaluation. Please refer to the appendix for more details on evaluation.

\begin{figure*}
    \centering
    \includegraphics[width=\linewidth]{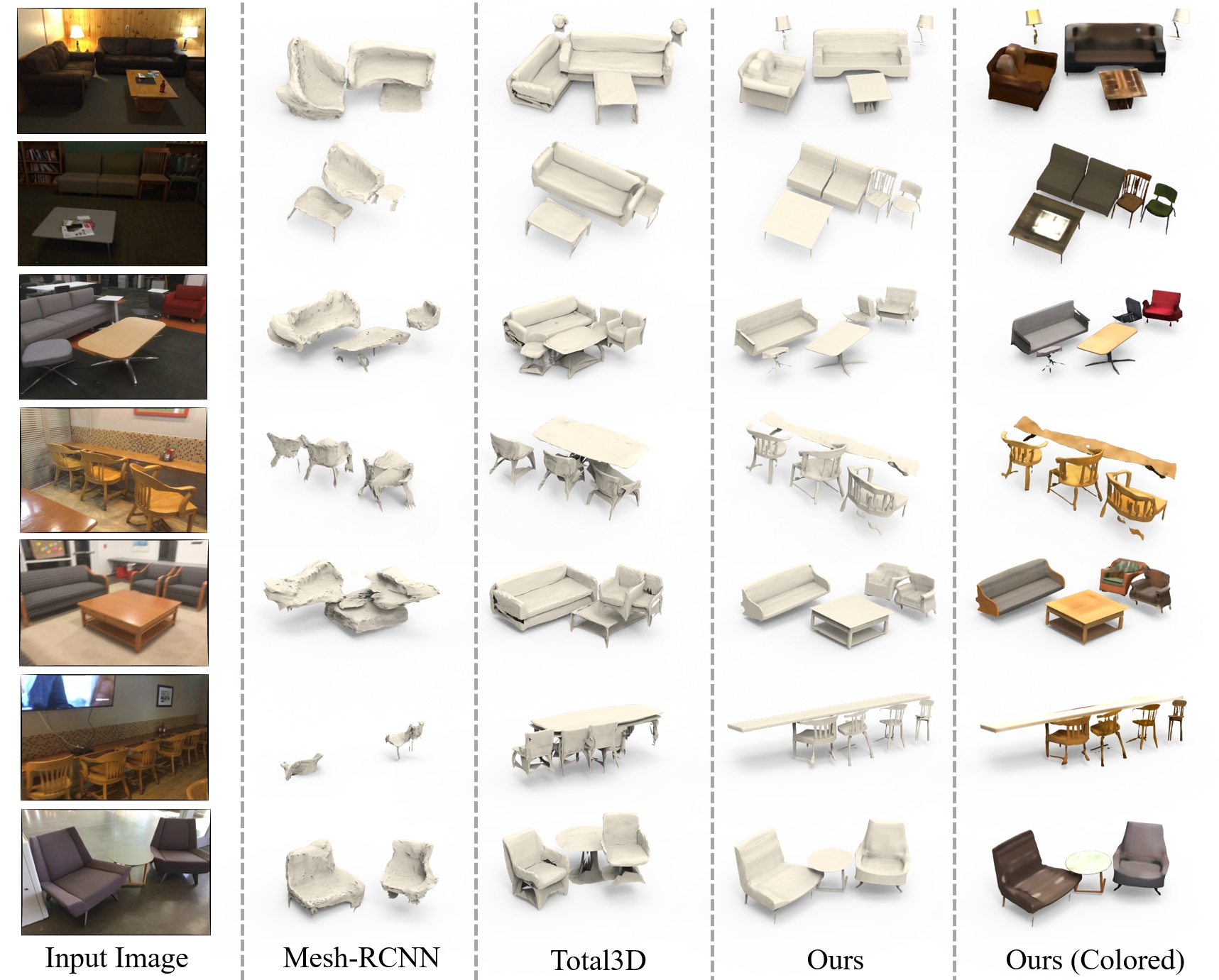}
    \vspace{-0.4cm}
    \caption{Comparisons of the scene reconstructions under the real-captured images from ScanNet.}
    \label{fig:scannet}
    \vspace{-0.4cm}
\end{figure*}

\subsection{Scene Reconstruction on 3D-Front}
Tab.~\ref{table:main} reports numerical comparisons on the 3D-Front dataset. We achieve the best performance among the state-of-the-art methods. Specifically, PanoRecon is trained under the 3D-Front dataset, therefore it shows convincing results in this dataset. Mesh R-CNN and Total3D are trained under Pix3D \cite{sun2018pix3d}/ShapeNet \cite{chang2015shapenet} and SUN-RGBD \cite{song2015sun}/Pix3D datasets, respectively. 

The qualitative comparison is shown in Fig.~\ref{fig:3dfront}, where we remove the background geometries for PanoRecon, ours and GT for a clear visual comparison on the generated instances among all the methods. We further show the colored scene since the used object generator Shap$\cdot$E is able to generate textured 3D objects. The visualization demonstrates our superior performance to produce accurate and visual-appealing scene reconstruction from merely a single image in a zero-shot manner.

\subsection{Scene Generation of Open-World Images}
We further evaluate our method under the open-world images from the BlendSwap dataset and the indoor scene dataset Replica. The quantitative comparisons in these two datasets are shown in Tab. \ref{table:main}, where we achieve the best performance over all the baseline methods. Note that the performance of PanoRecon \cite{dahnert2021panoptic} largely degrades under open-world scene images compared to the performance under 3D-Front dataset. The reason is that PanoRecon fails to generalize to the out-of-distribution inputs and can only handle the specific image patterns in the trained 3D-Front dataset. 

The visual comparison is shown in Fig. \ref{fig:replica}, where we significantly outperform the previous works in the generation accuracy and completeness. Specifically, as shown in the $3$-rd and $5$-th row in Fig. \ref{fig:replica}, our method generates more accurate geometries for the table with thin legs and the chair with a complex back. While Mesh R-CNN and Total3D can only generate the coarse 3D shapes and also fail to estimate accurate layout.

\subsection{Scene Reconstruction from Real Images}

We further evaluate deep prior assembly under the ScanNet \cite{dai2017scannet} using real images. For a qualitative comparison with other methods, we select 7 scenes from the test set of ScanNet and sample one image from each scene as the input. We compare deep prior assembly with the state-of-the-art methods in scene reconstruction from single images, e.g., Mesh R-CNN \cite{gkioxari2019mesh} and Total3D \cite{nie2020total3dunderstanding}. We do not compare with PanoRecon \cite{dahnert2021panoptic} here since it fails to generalize to the out-of-distribution inputs and can only handle the specific image patterns in the trained 3D-Front dataset, as demonstrated in Fig. \ref{fig:replica}. 

We show the visual comparisons in Fig. \ref{fig:scannet}, where we successfully reconstruct scenes from real images and significantly outperform the previous works in the reconstruction accuracy and completeness. This demonstrates the huge potentials of the assembled deep priors in reconstructing real-world 3D scenes. Note that the real-world images are often blurry and corrupted when the camera doesn’t focus well, e.g., the blurry input image shown in the 5-th row in Fig.~\ref{fig:scannet}. While our proposed deep prior assembly can also handle these challenging situations due to the powerful and robust deep priors from the large vision models.

\subsection{Ablation Study}
\begin{wraptable}[8]{r}{0.5\textwidth}
\setlength{\tabcolsep}{1mm}
\centering
\vspace{-0.5cm}
\resizebox{\linewidth}{!}{
\begin{tabular}{l|c|c|c}
\toprule

Ablation & CDL1-S $\downarrow$ & CDL1 $\downarrow$ & F-Score$\uparrow$ \\ 
\midrule
W/o Stable-Diffusion & 0.128 & 0.125 & 67.22  \\
W/o 2D-Matching & 0.124 & 0.121 & 68.42  \\
W/o 3D-Matching & 0.199 & 0.168 & 56.08 \\
\midrule
Full     & \textbf{0.113} & \textbf{0.110} & \textbf{70.48}  \\
\bottomrule
\end{tabular}}
\vspace{-0.2cm}
\caption{Ablation studies on framework designs.}

\label{table:ablation1}
\vspace{-0.2cm}
\end{wraptable}
\textbf{Framework Design.}
To evaluate the major components in our methods, we conduct ablations under the Replica dataset \cite{straub2019replica} and report the results in Tab.~\ref{table:ablation1}. We first justify the effectiveness of introducing Stable-Diffusion for enhancing and inpainting images as shown in `W/o Stable-Diffusion', where we directly adopt the segmented instances for shape generation without leveraging Stable-Diffusion for enhancing and inpainting them. We then report the performance of removing the 2D or 3D matching constraints as shown in `W/o 2D-Matching' and `W/o 3D-Matching'. The ablation studies demonstrate the effect of each design by significantly improving the generation performance. Note that the pose/scale optimization is broken without 3D-Matching since the only 2D-Matching does not involve depth information.

\begin{wraptable}[7]{r}{0.5\textwidth}
\setlength{\tabcolsep}{1mm}
\centering
\vspace{-0.4cm}
\resizebox{\linewidth}{!}{
\begin{tabular}{c|c|c|c|c}
\toprule

Open-CLIP & RANSAC & CDL1-S $ \downarrow$  & CDL1  $ \downarrow$  & F-Score$\uparrow$ \\ 
\midrule
\Checkmark & \XSolidBrush & 0.121 & {0.118} & {69.28}  \\
\XSolidBrush & \Checkmark & 0.129 & 0.123 & {68.92}  \\
\midrule
\Checkmark & \Checkmark & \textbf{0.113} & \textbf{0.110} & \textbf{70.48}  \\
\bottomrule
\end{tabular}}
\vspace{-0.2cm}
\caption{Ablation studies on the effect of Open-CLIP filtering and RANSAC-like solution.}
\label{table:ablation2}
\vspace{-0.35cm}
\end{wraptable}
\noindent\textbf{Effect of Open-CLIP and RANSAC-like solution.} We further evaluate the effectiveness of filtering bad samples with Open-CLIP and the RANSAC-like solution for robust pose / scale optimization. The results is shown in Tab.~\ref{table:ablation2}, where both components improve the scene reconstruction accuracy.

\section{Conclusion}
We introduce deep prior assembly, a novel framework that assembles diverse deep priors from large models for scene reconstruction from single images in a zero-shot manner. This approach breaks down the task into several sub-tasks, each of which is handled by a deep prior. We do not rely on any 3D or 2D data-driven training, and provide the key solutions on layout estimation and occlusion parsing to make all deep priors work together robustly. We report analysis, numerical and visual comparisons to show remarkable performance over the latest methods.

\section{Acknowledgement}
This work was supported by National Key R\&D Program of China (2022YFC3800600), the National Natural Science Foundation of China (62272263, 62072268), and in part by Tsinghua-Kuaishou Institute of Future Media Data.

\bibliographystyle{ieee_fullname}
\bibliography{sample}

\clearpage
\appendix

\leftline{\Large{\textbf{Appendix}}}

\section{More Ablation Studies and Analysis}

\begin{figure}[b]
	\centering

		\begin{minipage}{0.45\textwidth}
\centering
\resizebox{\linewidth}{!}{
\begin{tabular}{l|c|c|c}
\toprule

Ablation & CDL1-S $\downarrow$ & CDL1 $\downarrow$ & F-Score$\uparrow$ \\ 
\midrule
One-2-3-45 & 0.123 & 0.122 & 67.98  \\
EVA-CLIP & \textbf{0.113} & 0.111 &  70.41 \\
MiDaS & 0.120 & 0.119 & 68.71 \\
\midrule
Ours     & \textbf{0.113} & \textbf{0.110} & \textbf{70.48}  \\
\bottomrule
\end{tabular}}
\tabcaption{Ablation studies on the sub-task alternatives.}

\label{table:ablation3}
		\end{minipage}
  		\begin{minipage}[h]{0.54\linewidth}
    \centering
    \resizebox{1\linewidth}{2cm}{
    \includegraphics[width=1\linewidth]{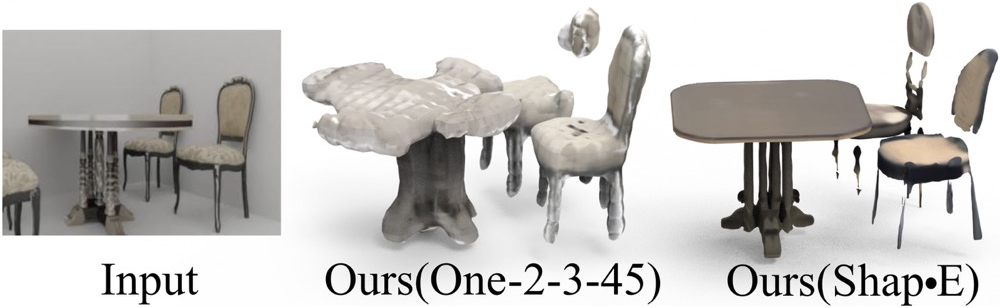}}

    \figcaption{Ablation on shape generation alternatives.}
    
    \label{fig:subtask}    
		\end{minipage}%
\end{figure}

\subsection{Alternatives on sub-tasks.} 
We explore the effectiveness of our chosen solutions in each sub-task by comparing them with the alternatives. Specifically, we conduct ablations to replace Shap$\cdot$E \cite{jun2023shap}  with One-2-3-45 \cite{liu2024one}, replace Open-CLIP \cite{radford2021learning, ilharco_gabriel_2021_5143773} with EVA-CLIP \cite{sun2023eva} and replace Omnidata \cite{eftekhar2021omnidata} with MiDaS \cite{Ranftl2022} in Tab.~\ref{table:ablation3}. We visually compare Shap$\cdot$E with One-2-3-45 for shape generation in Fig.~\ref{fig:subtask}, where the results demonstrate that Shap$\cdot$E is a more robust solution for generating 3D models from 2D instances.

\subsection{The Effect of Instance Scale}
\begin{figure*}[tb]
    \centering
    \includegraphics[width=\linewidth]{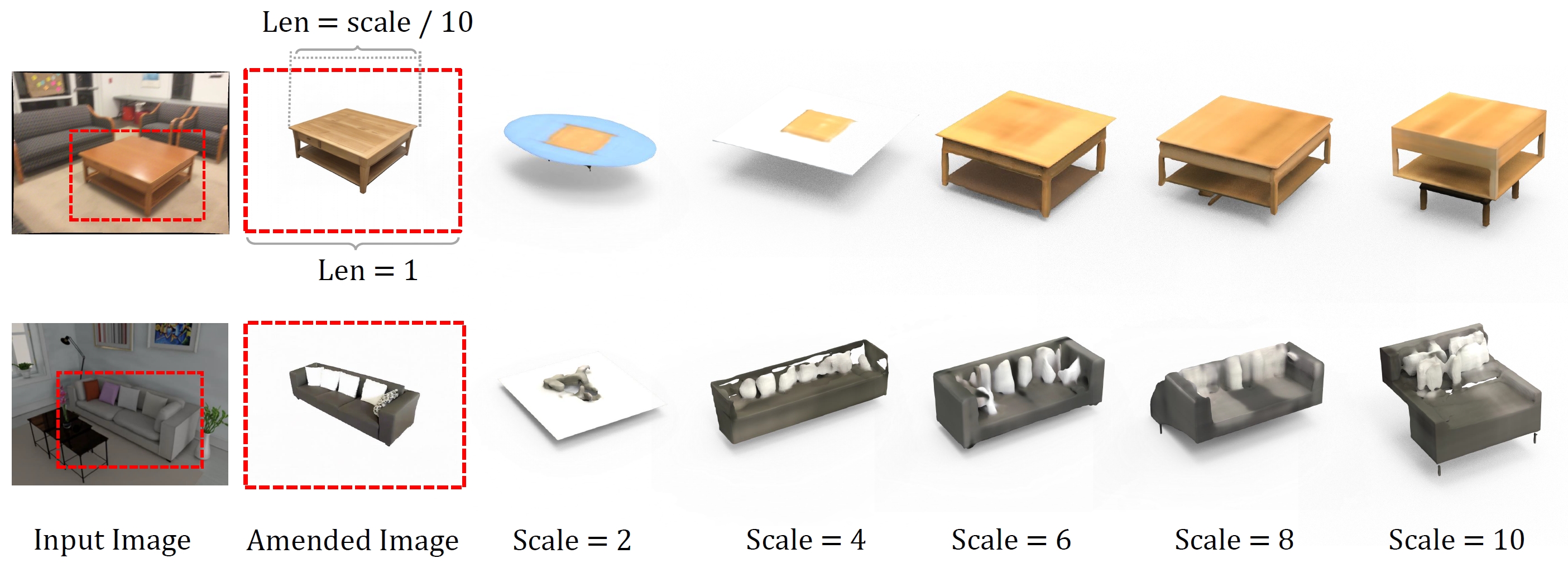}
    \caption{The ablation study on the instance scale. We select one instance for each input image and show the amended instance images. The generations obtained with Shap$\cdot$E under different instance scales are visualized on the right.}
    \label{fig:supp_scale}
\end{figure*}

We further conduct ablation studies to explore the effect of the instance scales to the generation qualities of Shap$\cdot$E as described in \textbf{``Enhance and Inpaint 2D instances''} of Sec.~3.2 in our paper. We provide an visual comparison of the generations with different instance scales in Fig. \ref{fig:supp_scale}. The results show that Shap$\cdot$E is quite sensitive to the scale of instances in the images, where a too small or too large scale will lead to inaccurate generations with unreliable geometries and appearances. We set the scale to 6 where the Shap$\cdot$E performs the best in shape generation from instance images according to our experiences.

\begin{figure}[tb]
    \centering
    \resizebox{0.65\linewidth}{!}{
    \includegraphics[width=1\linewidth]{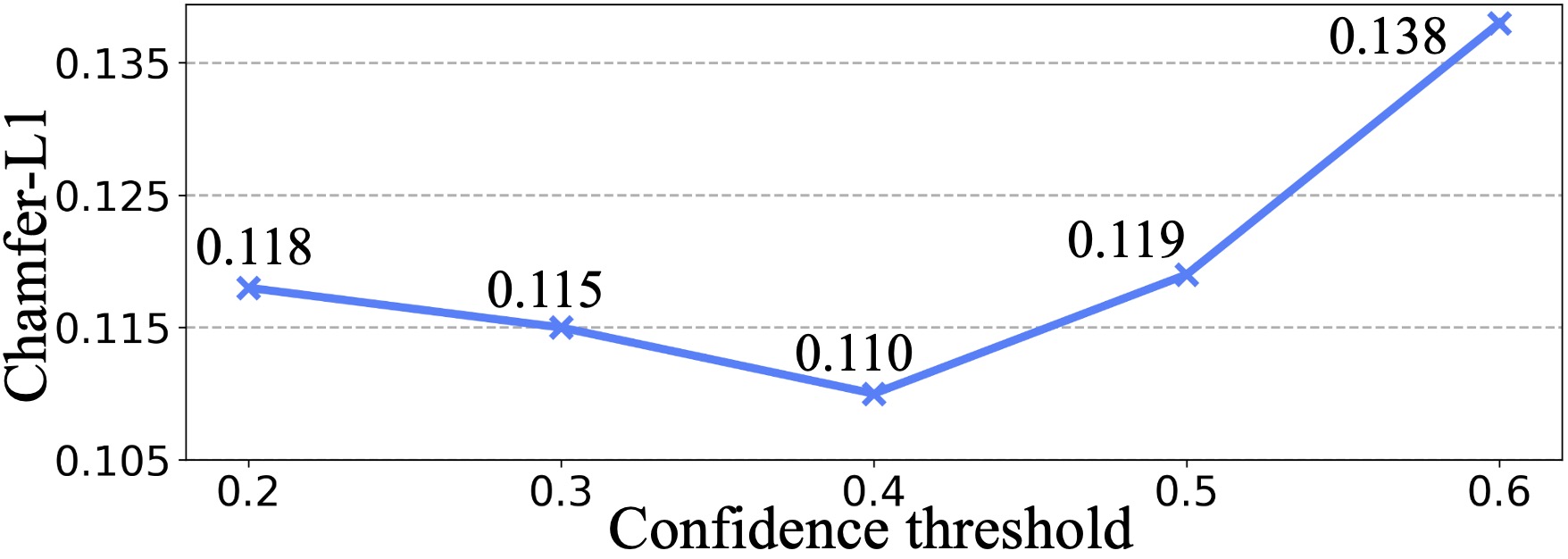}}
    \caption{The ablation study on the confidence threshold.}
    \label{fig:confidence_threshold}    
\end{figure}
\subsection{{The Effect of Confidence Threshold.}}
We further conduct ablations to evaluate the effect of confidence threshold $\sigma$ as described in \textbf{``Detect and Segment 2D instances.''} of Sec.3.2 in our paper. As shown in Fig.~\ref{fig:confidence_threshold}, a too large $\sigma$ will drop too many instances and a too small $\sigma$ struggles to filter bad instances. We choose $\sigma$ = 0.4 as the suitable confidence threshold.

\section{More Comparisons with Data-Driven Reconstruction Methods}

\begin{figure}[tb]
    \centering
    \resizebox{1\linewidth}{!}{
    \includegraphics[width=1\linewidth]{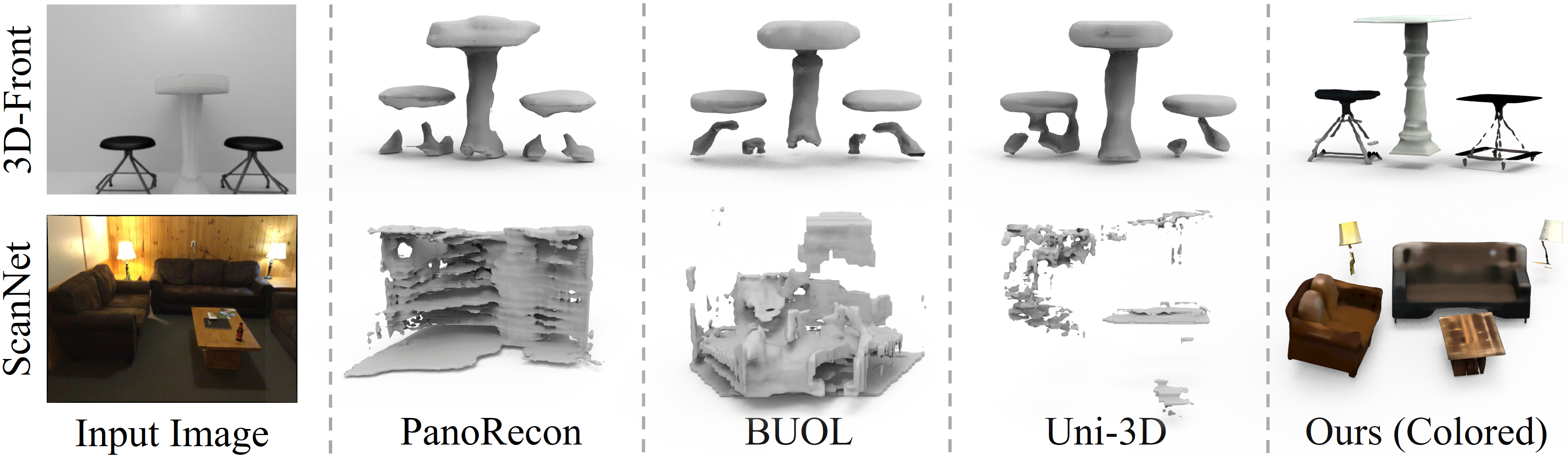}}
    \caption{Visual comparisons under 3D-Front and ScanNet dataset.}
    \label{fig:more_comp} 
\end{figure}

We additionally compare our method with SOTA data-driven scene reconstruction works PanoRecon \cite{dahnert2021panoptic}, BUOL \cite{chu2023buol} and Uni-3D \cite{zhang2023uni}. We show the visual comparisons under 3D-Front and ScanNet datasets in Fig.~\ref{fig:more_comp}, where our method achieves better and more visual-appealing results under both 3D-Front and ScanNet datasets. Specifically, our method significantly outperforms other methods using real-world images in ScanNet. The reason is that all the three methods are trained under 3D-Front, and struggle to generalize on real-world images. 

\begin{figure}[tb]
    \centering
    \includegraphics[width=0.7\linewidth]{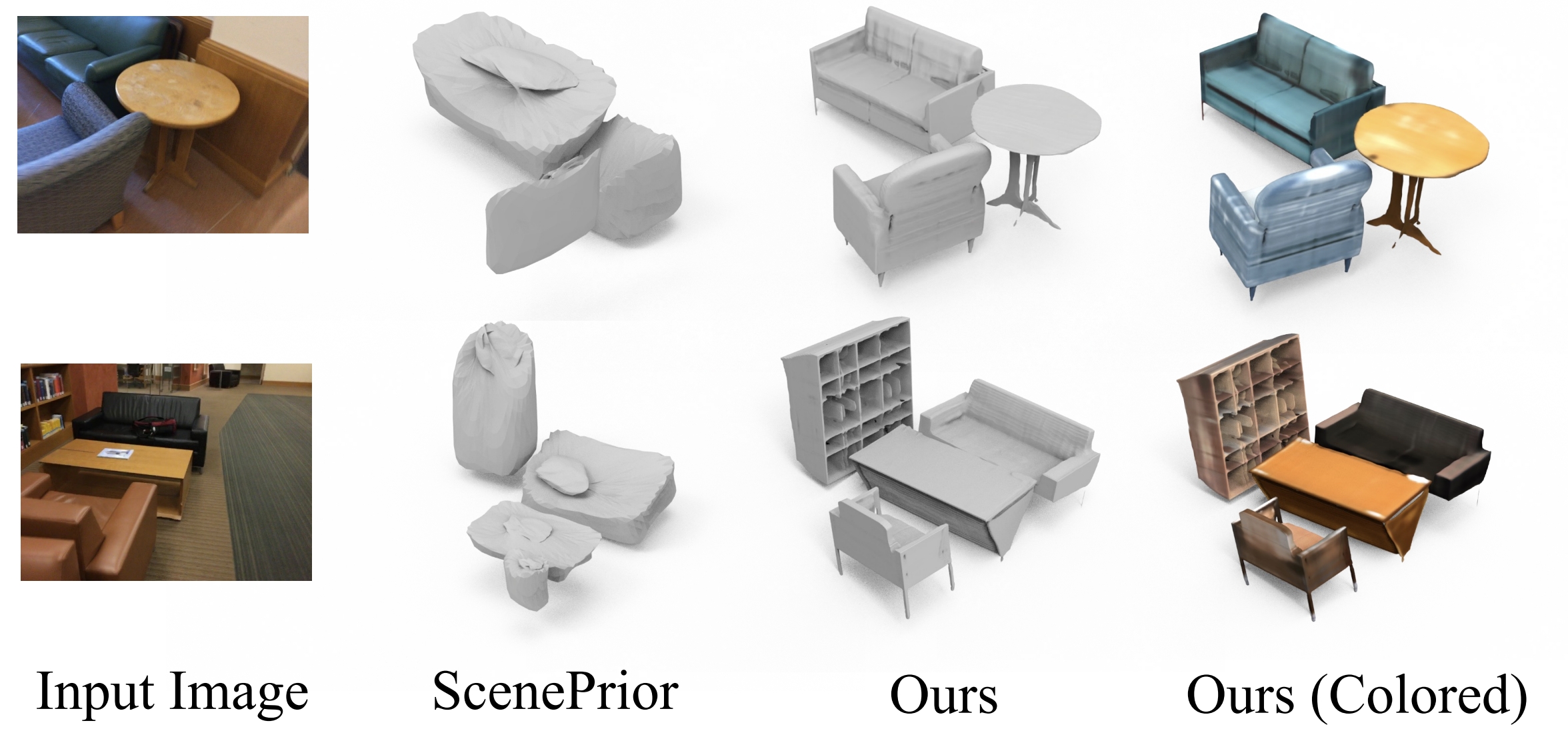}
    \caption{Visual comparisons with ScenePrior under ScanNet dataset.}
    \label{fig:sceneprior_comp} 
\end{figure}

We further compare deep prior assembly with ScenePrior \cite{nie2023learning} in Fig.\ref{fig:sceneprior_comp}. As shown, our method clearly outperforms ScenePrior in terms of the quality of scene geometries. Moreover, ScenePrior can only reconstruct the geometry, whereas deep prior assembly is capable of recovering high-fidelity scene appearances as well. 

\section{Background Reconstruction}
\begin{figure}[h]
    \centering
    \resizebox{1\linewidth}{!}{
    \includegraphics[width=1\linewidth]{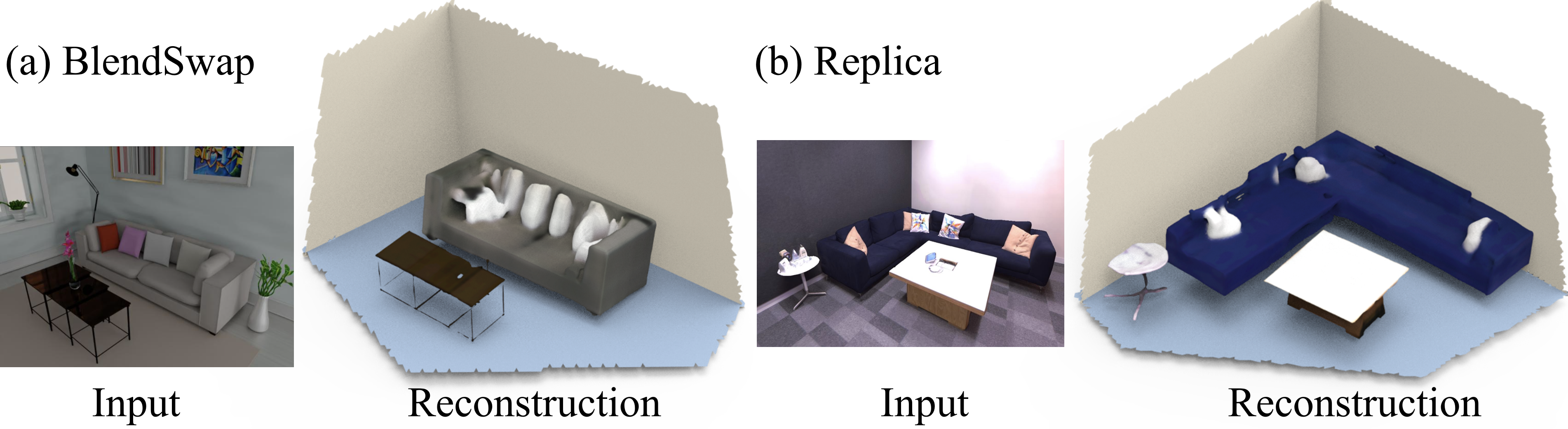}}
    \caption{Scene reconstructions with backgrounds.}
    \label{fig:background} 
    
\end{figure}

We demonstrate that deep prior assembly can also reconstruct the background geometries from the scene images. We show two scene reconstructions with backgrounds (i.e. floor, wall) in BlendSwap and Replica datasets in Fig.~\ref{fig:background}. The backgrounds are achieved by fitting planes to the projected background depth points in a similar way as our pose/scale optimization algorithm. We then cull the geometries outside of the view frustum for a clear visualization.

\section{Outdoor Scene Reconstruction}

\begin{figure}[tb]
    \centering
    \includegraphics[width=1\linewidth]{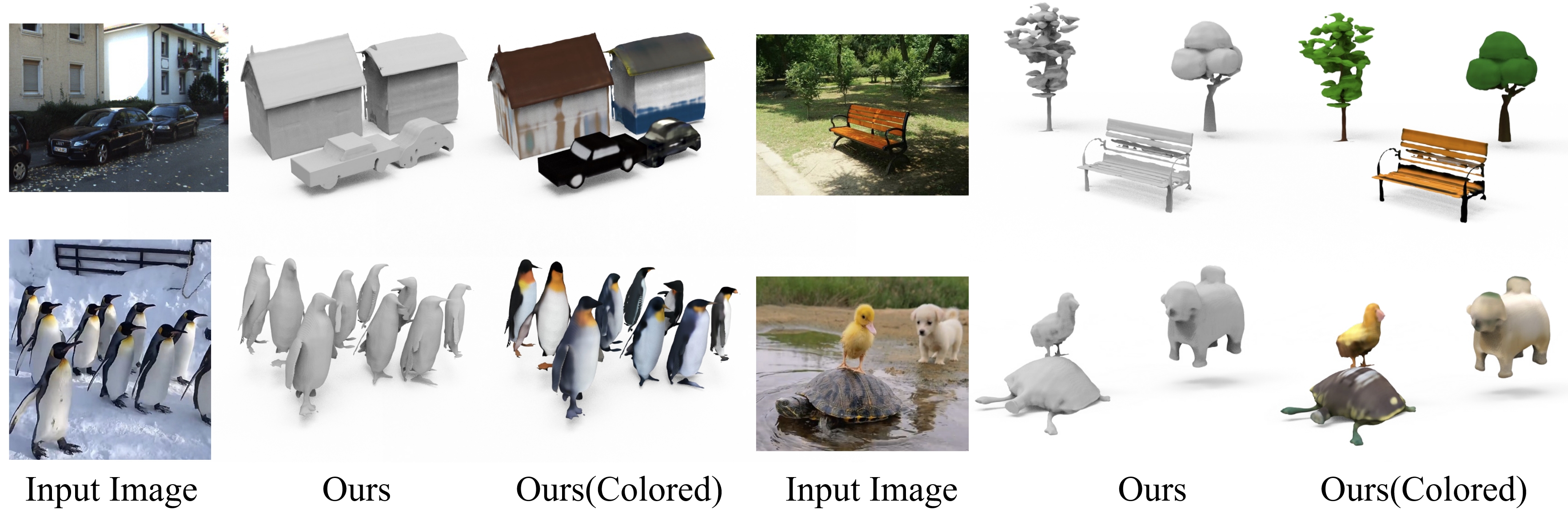}
    \caption{Outdoor scene reconstructions produced by deep prior assembly.}
    \label{fig:outdoor_scene} 
\end{figure}

We further conduct experiments to evaluate deep prior assembly on complex outdoor scenes and scene containing animals, as shown in Fig. \ref{fig:outdoor_scene}. The first image comes from KITTI dataset, others are collected from the Internet. With the help of powerful large foundation models, deep prior assembly demonstrates superior zero-shot scene reconstruction performance in these real-world outdoor scenes.

\section{Efficiency Analysis}

We further evaluate the efficiency of our proposed deep prior assembly by reporting the average runtime of each sub-pipeline in our framework. The results in Tab.~\ref{tab:runtime} show that reconstructing a scene from a single image takes less than 3 minutes in total, where the inference of Grounded-SAM, Open-CLIP and Omnidata takes only about 1 second. The most time consuming parts include the StableDiffusion, Shap$\cdot$E and the RANSAC-like pose/scale optimization. For these three parts, we process all instances of the scene in parallel, resulting in significant time savings.

\begin{table}[h]

\setlength{\tabcolsep}{2mm}

\resizebox{1.0\linewidth}{!}{
\begin{tabular}{c|c|c|c|c|c|c|c}
 
 \toprule
 ~ & G-SAM & Sta.-Diff. & Open-CLIP & Omnidata & Shap$\cdot$E & RANSAC-Opti & Total \\
 \midrule
 
 Time (s) & 0.93 & 33.6 & 0.05 & 0.21 & 39.2 & 97.2 & 171.2 \\
 \bottomrule
 \end{tabular}}
 \vspace{0.3cm}
 \caption{Runtime of each sub-pipeline on a RTX3090 GPU.}
 
 \label{tab:runtime}
 
\end{table}

\section{Evaluation Details}

\noindent\textbf{PanoRecon.} For evaluating PanoRecon \cite{dahnert2021panoptic}, we adopt the official code and pretrained models for inference and directly report the performance under 3D-Front \cite{fu20213d-front} dataset by evaluating the metrics (e.g. Chamfer Distance and F-Score) between the reconstructions and the ground truth meshes. For the Replica \cite{straub2019replica} and BlendSwap \cite{azinovic2022neural} datasets where the scene location and orientation do not match the 3D-Front dataset where the PanoRecon is trained, we first normalize the center of the predicted scenes to the ground truth scenes and then register the predicted scene reconstructions to the ground truth scenes. Specifically, we first predict an initial alignment by a global registration algorithm based on feature matching \cite{rusu2009fast} with RANSAC and then leverage ICP (Iterative Closest Point) registration algorithm \cite{rusinkiewicz2001efficient} to obtain the fine registration based on the initial alignment. The metrics are reported with the registered reconstructions and the ground truth meshes.

\noindent\textbf{Total3D.} 
We leverage the official code and the pretrained models for predicting scene reconstructions with Total3D  \cite{nie2020total3dunderstanding}.
We evaluate Total3D under 3D-Front, Replica and BlendSwap with a similar way as we evaluating PanoRecon to first normalize the predicted scenes and register them to the ground truth ones before computing metrics. Total3D only predicts 3D objects and do not generate backgrounds (e.g. wall and floor). Therefore, we further report the single-direction chamfer distance from the generated objects to the ground truth meshes, i.e., Chamfer-L1 (S), to only evaluate the accuracy of the generated objects. Note that Total3D can generate the scene layout which can roughly represent the background, however, we find that Total3D generates layouts with large errors on all the three datasets. Therefore we do not keep the layout of Total3D for evaluation.

\noindent\textbf{Mesh R-CNN.} We adopt the official code and the pretrained models for predicting scene reconstructions with Mesh R-CNN \cite{gkioxari2019mesh}. We notice that Mesh R-CNN produces reconstructions with larger scales than the predictions of other methods and the ground truths. Therefore, we first normalize both the center and scale of the predicted scenes to the ground truth scenes and then register the predicted scene reconstructions to the ground truth scenes with a similar way as we evaluate Total3D.

\noindent\textbf{Deep Prior Assembly.} 
We evaluate our proposed deep prior assembly in the same way. We follow the same settings as we evaluate other baselines to first normalize the center of the predicted scenes to the ground truth scenes, and then register the predicted scenes to the ground truth scenes. The background points (e.g. wall and floor) of deep prior assembly are obtained by back-projecting the background depth maps, i.e., the areas where no instances exists.

\section{Limitation}

One limitation of our method is that it may sometimes produce reconstructions with 3D instance models that are not perfectly aligned with the 2D instance segmentations in the input images. For example, the left chair generation in the last row of Fig.~\ref{fig:scannet} exhibits a different color from the 2D instance of the input image. The reason is that we use Stable-Diffusion to enhance and inpaint the 2D instances, and then leverage Shap$\cdot$E to generate 3D reconstructions. This process can introduce some randomness in the generated textures. The randomness primarily affects the appearances, while the geometries remain accurate. However, we justify that most reconstructions can faithfully recover the consistent scene appearances and geometries from the input images. 
Another limitation of deep prior assembly is that scene reconstructions may exhibit distortion due to inaccurate depth scale and shift. This issue can be addressed by replacing Omnidata with recent advances in metric depth estimation methods \cite{yang2024depth,ke2024repurposing}.

\end{document}